# Performance metrics in machine learning: a scenario-based guide from the confusion matrix to calibration, regression and uncertainty

Mehmet Baygin[1,*], Sengul Dogan[2], Turker Tuncer[2]

[1] Erzurum Technical University, Erzurum, Turkey

mehmet.baygin@erzurum.edu.tr

[2] Firat University, Elazig, Turkey

sdogan@firat.edu.tr; turkertuncer@firat.edu.tr

* Corresponding author: Mehmet Baygin, mehmet.baygin@erzurum.edu.tr

**Abstract**

Background and purpose: Machine-learning and deep-learning studies are compared through performance metrics. The same predictions can yield very different numbers depending on the metric, the class mix, the decision threshold, the averaging rule, the evaluation unit and the software convention. This methodological tutorial explains these mechanisms with controlled scenarios whose inputs every reader can recompute.

Materials and methods: Twenty-two confusion-matrix metrics were organized in a prevalence-coded metric map. Fifteen controlled scenarios (S1–S15) covered binary and multi-class classification, ranking and probability quality, regression, interval estimation, aggregation, reverse calculation and software conventions. Every scenario fixed its inputs in released files and changed one factor at a time. All values were recomputed with independent library functions (scikit-learn, statsmodels, SciPy and mlscorecheck) or independent re-implementations.

Results: All 205 verification checks passed, 202 of them independent and 3 internal. An always-negative rule reached an accuracy of 0.990 at a prevalence of 0.010, with a balanced accuracy of 0.500. At fixed sensitivity and specificity, precision fell from 0.889 to 0.008 as prevalence fell from 0.5 to 0.001. Three probability variants shared a ROC-AUC of 0.825, but their calibration slopes were 2.08, 1.04 and 0.52. With 10 participants, a participant-majority accuracy of 0.700 had a 95% interval of 0.348–0.933. Fold-averaged scores were flagged as inconsistent only when they were tested as one pooled test set.

Conclusions: The scenarios isolate mechanisms; they do not estimate how often each mechanism occurs in published studies. A literature audit and a real-data demonstration are specified as protocols in an appendix and were not executed. The released inputs, scripts, goal-based metric table and reporting checklist support transparent metric choice and reporting.

## 1. Introduction

Machine-learning and deep-learning studies are ranked by numbers. A reported accuracy, F1 score or area under the curve is read as a property of the model. It is, however, also a property of the test set, the decision threshold, the averaging rule, the evaluation unit and the software that computed it. The same predictions can therefore support very different summaries. On a test set with 1% positive cases, a rule that never predicts the positive class reaches an accuracy of 0.990 and a balanced accuracy of only 0.500.

The data behind a performance estimate are determined by the evaluation design. Leakage between training and test sets can inflate performance. The same risk arises if segment-level and participant-level estimates are combined [1, 2]. These design issues are outside this guide's scope. The design is held fixed here. Metric interpretation and reporting are examined for a fixed set of predictions.

Complete metric reports are required by guidance for prediction models and image analysis [3, 4]. Yet the reasons behind these recommendations are not always clear. Accuracy is widely accepted as misleading under imbalance. The size of this effect depends on numerical inputs. These are rarely provided, so readers often cannot repeat the calculation.

This study is a methodological tutorial on performance metrics with released and independently verified inputs. Constructed scenarios are presented; an empirical literature survey was not performed. A prevalence-coded map is used to organize 22 confusion-matrix metrics. One mechanism is examined in each of 15 controlled scenarios. Classification, probability quality, regression and reporting are covered. All values are recomputed with independent software. Internal re-derivations are counted separately. Reported values can be screened with a consistency framework that separates inconsistent reports from reports that cannot be assessed. All proportions are reported on the 0–1 scale.

### 1.1. Literature review

Surveys of classification measures established the vocabulary used in this study. Sokolova and Lapalme organized measures by their invariance to changes in the confusion matrix [5]. Ferri and colleagues grouped measures through experimental comparisons across datasets [6]. Binary and multi-class assessment methods were reviewed by Tharwat [7]. Powers described the links between precision, recall, informedness, markedness and correlation [8].

Class imbalance is often cited as a source of misleading metric values. Its effects were quantified by Luque and colleagues [9]. Brodersen and colleagues recommended balanced accuracy for unequal classes [10]. All four binary confusion-matrix cells are used in MCC [11]. Chicco and Jurman preferred MCC to accuracy and F1 for binary evaluation [12]. Gorodkin extended it to K categories [13]. Sensitivity and specificity are combined in Youden's index [14].

Paradoxes also occur in agreement statistics. Agreement expected by chance is accounted for in Cohen's kappa [15]. Feinstein and Cicchetti described high agreement with low kappa under skewed marginal totals [16].

Threshold-free summaries measure different aspects of performance. The true-positive and false-positive rate trade-off is described by ROC analysis [17]. ROC-AUC represents the probability that a random positive case scores above a random negative case [18]. Hand identified inconsistent misclassification costs across classifiers in this measure [19]. Precision–recall plots were preferred to ROC plots for imbalanced data by Saito and Rehmsmeier [20]. Davis and Goadrich showed that linear interpolation between precision–recall points is not achievable [21].

Probability quality is a separate property from discrimination. The Brier score and the logarithmic score are proper scoring rules [22, 23]. Guo and colleagues reported that modern neural networks are often overconfident [24]. The expected calibration error became the common summary of this effect [25]. Van Calster and colleagues described calibration as the weak point of predictive analytics [26]. Steyerberg and colleagues placed calibration and discrimination in one framework [27], and decision curve analysis adds a clinical-utility view [28].

Regression metric choice has also been debated. Willmott and Matsuura preferred mean absolute error to root mean square error [29]. Root mean square error was defended for Gaussian errors by Chai and Draxler [30]. Instability of percentage errors near zero was demonstrated by Hyndman and Koehler [31]. Chicco and colleagues compared the coefficient of determination with percentage and absolute errors [32].

Uncertainty and model comparisons require inferential methods. Wilson and Clopper–Pearson intervals are used for binomial proportions [33, 34]. Their coverage was compared by Brown and colleagues [35]. Paired classifiers on one test set can be compared with McNemar's test [36]. DeLong's method is used for correlated ROC curves [37]. Dietterich examined test error rates [38]. Bootstrap intervals cover other metrics [39]. Cluster-level resamples are required for clustered data [40].

Domain guidance and tools address several of these issues. Hicks and colleagues explained metric interpretation in medical artificial intelligence [41]. They also released MediMetrics. Suitable metrics were linked to image-analysis problems in Metrics Reloaded [4]. Reinke and colleagues catalogued metric-related pitfalls [42]. Required prediction-model report items are specified in TRIPOD+AI [3].

Reported values can be checked for mathematical consistency. In the GRIM test, a reported mean is checked against the stated sample size [43]. Consistency tests for binary classification scores were developed by Fazekas and Kovács [44]. These tests were released in mlscorecheck by Kovács and Fazekas [45]. Cross-validation, multi-class classification and regression are also supported by the package.

### 1.2. Literature gaps

Four gaps concern access to inputs, coverage across metric families, software conventions and interpretation of consistency checks.

- Metric behaviour is usually explained in words or through small examples. Inputs are often unavailable. Readers can follow the argument but cannot repeat its calculation or vary one factor.
- Threshold-based, probabilistic and regression metrics are usually discussed separately. A single fixed set of scenarios across these families is rarely provided with common notation.
- Undefined cases and software conventions are described in documentation. Their numerical effects are seldom quantified. The effect on a reported value is therefore difficult to assess.
- Consistency tests are available. Yet the effects of numerical precision, known class counts and the aggregation model are rarely explained. A failed check may be read as evidence of an error despite insufficient information for assessment.

### 1.3. Motivation and proposed guide

Each metric mechanism is illustrated with values traceable to fixed inputs. A metric map separates prevalence-dependent and prevalence-independent metrics at fixed sensitivity and specificity. One factor is varied in each of 15 scenarios. A common template is used: purpose, locked inputs, manipulated factor, results, figure, interpretation and reporting recommendation. Values are computed from formulas and checked with independent libraries. Consistency checks assign reported values to four outcome classes. Application goals are linked to primary and complementary metrics. Figure 1 summarizes the study structure.

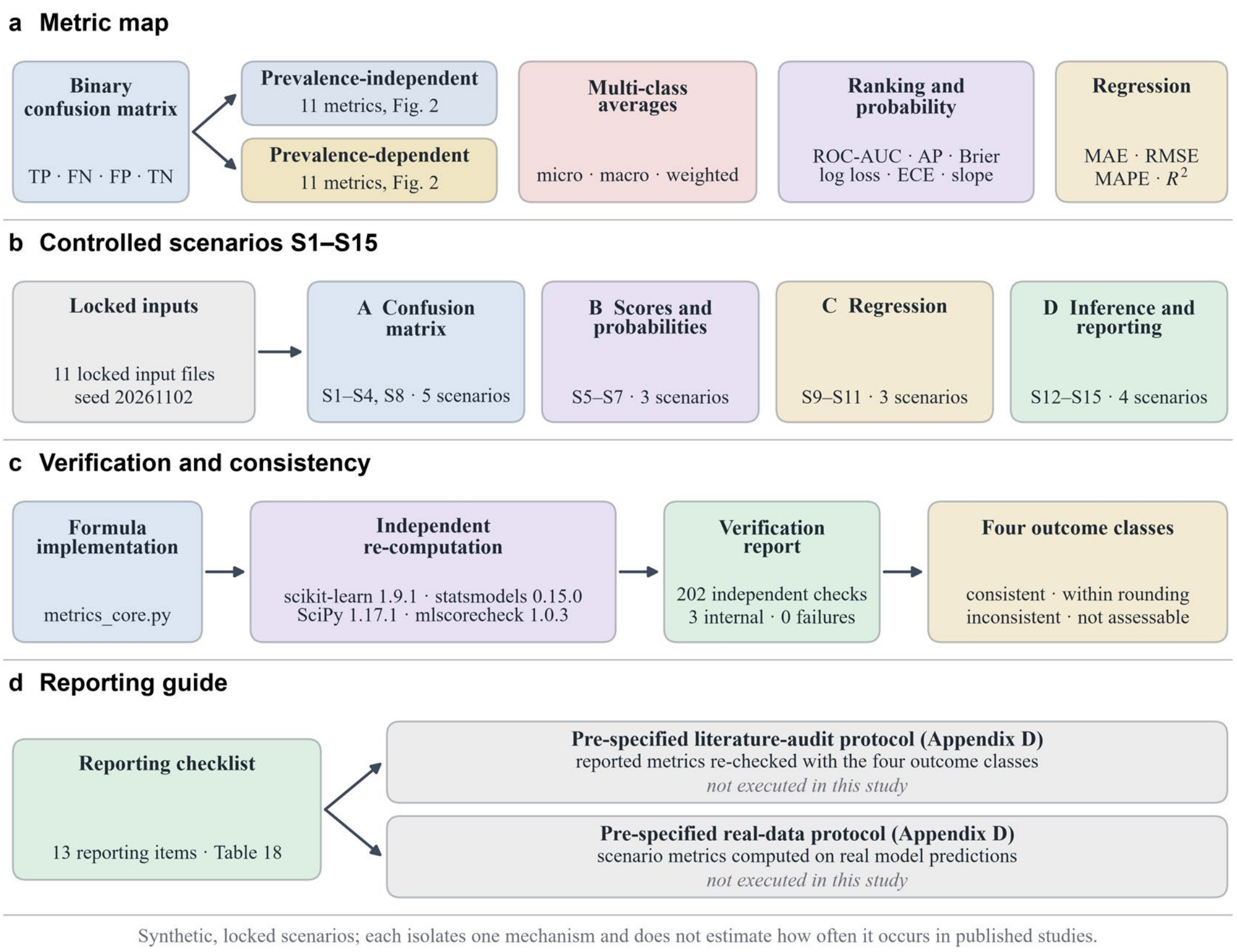

Fig. 1. Structure of the scenario-based guide to performance metrics. (a) Metric map: the binary confusion matrix yields 11 prevalence-independent and 11 prevalence-dependent metrics (Figure 2), beside multi-class averaging rules, ranking and probability metrics and regression metrics; (b) controlled scenarios S1–S15 in four blocks (A confusion matrix, B scores and probabilities, C regression, D inference and reporting), built from 11 locked input files with base seed 20261102; (c) verification and consistency: the formula implementation is re-computed with scikit-learn 1.9.1, statsmodels, SciPy and mlscorecheck at the versions shown (202 independent and 3 internal checks, 0 failures), and each consistency check is assigned to one of four outcome classes; (d) reporting guide: a checklist of 13 reporting items (Table 18) linked to scenarios S1–S15, and two pre-specified protocols in Appendix D (literature audit and real-data demonstration). Neither protocol was executed. Stages within each band are grouped by card colour. The colours from Figure 2 are used for the two prevalence groups in band a. Workflow order is shown by arrows. The scope of interpretation for the synthetic scenarios is stated in the grey footer.

The metric map, scenarios, verification layer and reporting guide are linked in Figure 1. The organization of the guide is shown. No numerical estimate is presented.

### 1.4. Novelties and contributions

The main contributions are listed below.

- Fifteen controlled scenarios are provided with released inputs. One mechanism is isolated in each scenario. All values can be recomputed from the files.
- Common notation is used for confusion-matrix, ranking, probability, regression and inferential metrics. The metric families are linked by a prevalence-coded map. Their use in applications is described in a goal-based table.
- Undefined cases and software conventions are quantified with identical inputs. The effect of each option on the reported value is demonstrated.
- A consistency framework with four outcome classes is presented and demonstrated with mlscorecheck. Inconsistent reports are distinguished from reports with insufficient information for assessment. Numerical precision is stated for every case.

The contributions concern metric interpretation and reporting. No new metric is proposed. The frequency of these mechanisms in published studies is not estimated.

### 1.5. Novelty boundary relative to established evaluation methods

The new elements are the fixed scenario set, verification code and prevalence-coded metric map. A four-class scheme is also provided for reports of consistency checks. All metrics, intervals and tests are established methods [5, 35]. The consistency tests of Fazekas and Kovács are applied through mlscorecheck [44, 45]. Existing metric calculators include MediMetrics [41]. The scenarios are synthetic. Each was designed to isolate a mechanism. Effect sizes in real studies cannot be estimated from them. A literature audit is specified in Appendix D for this purpose. It was not executed.

## 2. Materials and methods

The notation, metric families and scenario design are defined in this section. Uncertainty estimation, consistency checks and verification procedures are then described. The scope of the study is also stated.

### 2.1. Notation and the confusion matrix

A binary test set holds $P$ actual positive and $N$ actual negative cases, with $T = P + N$. At a fixed decision threshold, the classifier produces $TP$ true positives, $FN$ false negatives, $FP$ false positives and $TN$ true negatives. The predicted totals are $PP = TP + FP$ and $PN = FN + TN$. The prevalence is $\pi = P/T$. The positive class is specified in every scenario. Several metric values are altered by a swap of class labels. The 22 metrics derived from these counts are mapped in Figure 2.

| | | | | | |
|---|---|---|---|---|---|
| Total<br>$T = P + N$<br>**1,000** | Predicted positive<br>$PP = TP + FP$<br>**170** | Predicted negative<br>$PN = FN + TN$<br>**830** | Prevalence<br>$\pi = P / T$<br>**0.100** | Accuracy<br>$Acc = \frac{TP+TN}{T}$<br>**0.890** | Balanced accuracy<br>$BA = \frac{TPR+TNR}{2}$<br>**0.850** |
| Actual positive<br>$P = TP + FN$<br>**100** | True positive<br>$TP$<br>**80** | False negative<br>$FN$<br>**20** | Sensitivity, recall<br>$TPR = \frac{TP}{P}$<br>**0.800** | Miss rate<br>$FNR = \frac{FN}{P}$<br>**0.200** | Likelihood ratio (+)<br>$LR^+ = \frac{TPR}{FPR}$<br>**8.00** |
| Actual negative<br>$N = FP + TN$<br>**900** | False positive<br>$FP$<br>**90** | True negative<br>$TN$<br>**810** | False alarm rate<br>$FPR = \frac{FP}{N}$<br>**0.100** | Specificity<br>$TNR = \frac{TN}{N}$<br>**0.900** | Likelihood ratio (−)<br>$LR^- = \frac{FNR}{TNR}$<br>**0.222** |
| Counts<br>Unchanged at fixed TPR and TNR | Precision<br>$PPV = \frac{TP}{PP}$<br>**0.471** | False omission rate<br>$FOR = \frac{FN}{PN}$<br>**0.024** | Informedness<br>$BM = TPR + TNR - 1$<br>**0.700** | Markedness<br>$MK = PPV + NPV - 1$<br>**0.446** | Diagnostic odds ratio<br>$DOR = \frac{LR^+}{LR^-}$<br>**36.0** |
| Changes with prevalence $\pi$ | False discovery rate<br>$FDR = \frac{FP}{PP}$<br>**0.529** | Neg. predictive value<br>$NPV = \frac{TN}{PN}$<br>**0.976** | $F_1$ score<br>$F_1 = \frac{2TP}{2TP+FP+FN}$<br>**0.593** | Matthews correlation coefficient<br>$MCC = \frac{TP \cdot TN - FP \cdot FN}{\sqrt{PP \cdot PN \cdot P \cdot N}}$<br>**0.559** | |
| Worked example:<br>$TP = 80$, $FN = 20$<br>$FP = 90$, $TN = 810$ | Jaccard index (CSI)<br>$\frac{TP}{TP+FP+FN}$<br>**0.421** | Fowlkes–Mallows<br>$\sqrt{PPV \cdot TPR}$<br>**0.614** | Cohen's kappa<br>$\kappa = \frac{Acc - p_e}{1 - p_e}$<br>**0.534** | Geometric mean<br>$\sqrt{TPR \cdot TNR}$<br>**0.849** | Prevalence threshold<br>$\frac{\sqrt{TPR \cdot FPR} - FPR}{TPR - FPR}$<br>**0.261** |

Fig. 2. Map of confusion-matrix metrics with a numerical example. Each cell contains a metric name, formula and value. The example uses TP = 80, FN = 20, FP = 90 and TN = 810, at a prevalence of 0.100. Blue cells mark metrics unchanged at fixed TPR and TNR. Gold cells mark metrics affected by prevalence π. Counts and margins are shown in grey cells. Proportions are reported on the 0–1 scale to three decimals. LR+ is reported to two decimals and DOR to one. Counts in the confusion-matrix cells are coloured blue (true) and red (false). In the kappa cell, $p_e = (P \cdot PP + N \cdot PN) / T^2$ is the chance agreement.

Figure 2 shows each metric with its formula and its value for one worked confusion matrix. Its colour coding states which metrics stay fixed when only prevalence changes and does not rank the metrics.

The colour coding was verified numerically. A second matrix with the same sensitivity and specificity but a prevalence of 0.010 left all 11 blue-coded metrics unchanged and changed all 11 gold-coded metrics. Section 3.1 examines this mechanism over six prevalences.

## 2.2. Metric families

### 2.2.1. Threshold-based binary metrics

Accuracy is the proportion of correct decisions. It is defined in Eq. (1).

$$\mathrm{Acc} = \frac{TP + TN}{T} = \pi\,\mathrm{TPR} + (1 - \pi)\,\mathrm{TNR}, \quad (1)$$

where TPR is the true-positive rate (sensitivity) and TNR is the true-negative rate (specificity). In Eq. (1), the second form shows that accuracy is a prevalence-weighted mean of the two class-wise rates. This identity is used again in Section 2.5.

The class-wise rates condition on the actual class, whereas the predictive values condition on the predicted class. Eq. (2) defines the two pairs.

$$\mathrm{TPR} = \frac{TP}{P}, \quad \mathrm{TNR} = \frac{TN}{N}, \quad \mathrm{PPV} = \frac{TP}{PP}, \quad \mathrm{NPV} = \frac{TN}{PN}, \tag{2}$$

where PPV is the positive predictive value (precision) and NPV is the negative predictive value. In Eq. (2), $PP$ and $PN$ are the predicted positive and negative totals. The rates in the first pair do not depend on $\pi$; the predictive values do.

The $F_\beta$ score combines precision and recall with a weight on recall. Eq. (3) gives its count form.

$$F_\beta = \frac{(1+\beta^2)\,TP}{(1+\beta^2)\,TP + \beta^2 FN + FP}, \tag{3}$$

where $\beta > 0$ sets the relative weight of recall. In Eq. (3), $\beta = 1$ gives the F1 score, $2TP/(2TP + FP + FN)$. The score does not contain $TN$, which Section 3.1 examines.

Balanced accuracy and informedness summarize the two class-wise rates. Eq. (4) defines them.

$$\mathrm{BA} = \frac{\mathrm{TPR} + \mathrm{TNR}}{2}, \quad \mathrm{BM} = \mathrm{TPR} + \mathrm{TNR} - 1, \tag{4}$$

where BA is balanced accuracy and BM is bookmaker informedness, which equals Youden's index [14]. In Eq. (4), both quantities use only the class-wise rates and are therefore unchanged by prevalence.

The Matthews correlation coefficient is the Pearson correlation between actual and predicted binary labels [11]. Eq. (5) gives it in marginal form.

$$\mathrm{MCC} = \frac{TP \cdot TN - FP \cdot FN}{\sqrt{PP \cdot PN \cdot P \cdot N}}, \tag{5}$$

where the denominator is the product of the four marginal totals. In Eq. (5), MCC is undefined when any marginal total is zero. Swapping the class labels leaves it unchanged.

Cohen's kappa compares observed agreement with agreement expected from the marginal totals [15]. Eq. (6) defines it.

$$\kappa = \frac{\mathrm{Acc} - p_e}{1 - p_e}, \quad p_e = \frac{P \cdot PP + N \cdot PN}{T^2}, \tag{6}$$

where $p_e$ is the chance agreement implied by the marginal totals. In Eq. (6), the observed agreement is accuracy. The remaining metrics of Figure 2 follow from the same counts; Table 1 lists their ranges, undefined cases and infinite cases.

Table 1. Properties of the metrics used in this guide. Prevalence: whether the value changes when only the class mix changes at fixed class-wise rates (for score metrics, at fixed class-conditional score distributions); not applicable for regression. Undefined: the formula is 0/0 or has no value. Infinite: a non-zero numerator meets a zero denominator, or a logarithm of zero occurs. LR+ and LR− = positive and negative likelihood ratios; DOR = diagnostic odds ratio; MK = markedness; FM = Fowlkes–Mallows index; PT = prevalence threshold.

| Metric | Range | Best | Prevalence | Undefined (0/0) | Infinite |
|---|---|---|---|---|---|
| Accuracy | 0 to 1 | 1 | changes | T = 0 | never |
| Sensitivity (TPR) | 0 to 1 | 1 | fixed | P = 0 | never |
| Specificity (TNR) | 0 to 1 | 1 | fixed | N = 0 | never |
| Precision (PPV) | 0 to 1 | 1 | changes | PP = 0 | never |
| NPV | 0 to 1 | 1 | changes | PN = 0 | never |
| F1 | 0 to 1 | 1 | changes | TP + FP + FN = 0 | never |
| Balanced accuracy | 0 to 1 | 1 | fixed | P = 0 or N = 0 | never |
| Informedness (BM) | −1 to 1 | 1 | fixed | P = 0 or N = 0 | never |
| MK | −1 to 1 | 1 | changes | PP = 0 or PN = 0 | never |
| MCC | −1 to 1 | 1 | changes | any marginal total = 0 | never |
| Cohen’s kappa | at most 1 | 1 | changes | chance agreement = 1 | never |
| LR+ | 0 to ∞ | higher | fixed | TPR = FPR = 0 | FPR = 0 and TPR > 0 |
| LR− | 0 to ∞ | lower | fixed | FNR = TNR = 0 | TNR = 0 and FNR > 0 |
| DOR | 0 to ∞ | higher | fixed | TP · TN = FP · FN = 0 | FP · FN = 0 and TP · TN > 0 |
| Jaccard index | 0 to 1 | 1 | changes | TP + FP + FN = 0 | never |
| FM | 0 to 1 | 1 | changes | PP = 0 or P = 0 | never |
| Geometric mean | 0 to 1 | 1 | fixed | P = 0 or N = 0 | never |
| PT | 0 to 1 | context-dependent | fixed | TPR = FPR | never |
| ROC-AUC | 0 to 1 | 1 | fixed | P = 0 or N = 0 | never |
| Average precision | 0 to 1 | 1 | changes | P = 0 | never |
| Brier score | 0 to 1 | 0 | changes | T = 0 | never |
| Log loss | 0 to ∞ | 0 | changes | T = 0 | p = 0 or 1 for the wrong label |
| ECE | 0 to 1 | 0 | changes | T = 0 | never; value depends on the bins |
| MAE, RMSE, MedAE | 0 to ∞ | 0 | not applicable | T = 0 | never |
| MAPE | 0 to ∞ | 0 | not applicable | target = error = 0 | target = 0 and error ≠ 0 |
| sMAPE | 0 to 2 | 0 | not applicable | target = prediction = 0 | never |
| $R^2$ | −∞ to 1 | 1 | not applicable | constant targets with zero error | constant targets with non-zero error |

Table 1 fixes the properties that the scenarios test one by one. Its prevalence column refers to a change in class mix at fixed class-wise rates and does not describe other kinds of distribution shift.

### 2.2.2. Multi-class averaging

A K-class confusion matrix gives one-versus-rest counts $TP_k$, $FP_k$ and $FN_k$ for each class $k$ with support $n_k$. Eq. (7) defines the three common F1 summaries.

$$F_1^{\text{macro}} = \frac{1}{K}\sum_{k=1}^{K} F_{1,k}, \quad F_1^{\text{weighted}} = \sum_{k=1}^{K} \frac{n_k}{T} F_{1,k}, \quad F_1^{\text{micro}} = \frac{2\sum_k TP_k}{2\sum_k TP_k + \sum_k FP_k + \sum_k FN_k}, \tag{7}$$

where $F_{1,k}$ is the F1 score of class $k$ against all other classes. In Eq. (7), $T$ is the number of cases. In single-label problems the micro average equals accuracy, and the macro average of recall equals balanced accuracy. The multi-class MCC is Gorodkin's $R_K$ [13].

### 2.2.3. Ranking and probability metrics

The ROC-AUC is computed from scores without a threshold. Eq. (8) gives its Mann–Whitney form [18].

$$\text{AUC} = \frac{1}{PN_{\text{pairs}}} \sum_{i\in\mathcal{P}} \sum_{j\in\mathcal{N}} \left[\mathbb{1}(s_i > s_j) + 1/2\,\mathbb{1}(s_i = s_j)\right], \tag{8}$$

where $s_i$ is the score of case $i$, $\mathcal{P}$ and $\mathcal{N}$ are the sets of positive and negative cases, and $PN_{\text{pairs}} = P \cdot N$. In Eq. (8), $\mathbb{1}(\cdot)$ is the indicator function and ties count one half. Average precision summarizes the precision–recall curve as a step function. Eq. (9) defines it.

$$\text{AP} = \sum_{m} (R_m - R_{m-1}) P_m, \tag{9}$$

where $P_m$ and $R_m$ are precision and recall at the $m$-th distinct score threshold, ordered by decreasing score, and $R_0 = 0$. In Eq. (9), a non-informative ranking has a population precision equal to the prevalence $\pi$ at every recall level, so $\pi$ is the population reference line [20]. In a finite sample the expected AP of a random ranking exceeds $\pi$. With one positive and one negative case, the two possible rankings give AP values of 1 and 0.5, so the expected AP is 0.75 at a prevalence of 0.5.

Probability quality is summarized by proper scoring rules [23]. Eq. (10) defines the Brier score and the log loss.

$$\text{Brier} = \frac{1}{T}\sum_{i=1}^{T}(p_i - y_i)^2, \quad \text{LogLoss} = -\frac{1}{T}\sum_{i=1}^{T}[y_i \ln p_i + (1-y_i)\ln(1-p_i)], \tag{10}$$

where $p_i$ is the predicted probability of the positive class and $y_i \in \{0,1\}$ is the label. In Eq. (10), lower values are better for both rules [22]. The expected calibration error bins the probabilities. Eq. (11) defines it.

$$\mathrm{ECE}_B = \sum_{b=1}^{B} \frac{n_b}{T} \left| \bar{y}_b - \bar{p}_b \right|, \tag{11}$$

where $B$ is the number of equal-width bins, $n_b$ is the number of cases in bin $b$, $\bar{y}_b$ is their observed positive rate and $\bar{p}_b$ is their mean predicted probability [25]. In Eq. (11), empty bins contribute zero. The calibration slope and intercept come from logistic models on the logit of the predicted probability [26]. Eq. (12) states the two models.

$$\operatorname{logit} \Pr(y_i = 1) = a + b \operatorname{logit}(p_i), \qquad \operatorname{logit} \Pr(y_i = 1) = a_0 + \operatorname{logit}(p_i), \tag{12}$$

where $b$ is the calibration slope and $a_0$ is the calibration intercept with the slope fixed at one. In Eq. (12), a well-calibrated model has $b = 1$ and $a_0 = 0$. A slope below one indicates overconfident probabilities, and a slope above one indicates underconfident probabilities.

#### 2.2.4. Regression metrics

Regression errors are $e_i = \hat{y}_i - y_i$ for $T$ test targets. Eq. (13) defines the two absolute-scale summaries.

$$\mathrm{MAE} = \frac{1}{T} \sum_{i=1}^{T} |\, e_i|, \quad \mathrm{RMSE} = \sqrt{\frac{1}{T} \sum_{i=1}^{T} e_i^2}, \tag{13}$$

where $\hat{y}_i$ is the prediction and $y_i$ is the target. In Eq. (13), RMSE $\geq$ MAE always holds for one error vector with equal weights. The median absolute error (MedAE) replaces the mean by the median. Eq. (14) defines the relative errors.

$$\mathrm{MAPE} = \frac{1}{T} \sum_{i=1}^{T} \frac{|e_i|}{|y_i|}, \quad \mathrm{sMAPE} = \frac{1}{T} \sum_{i=1}^{T} \frac{2|e_i|}{|y_i| + |\hat{y}_i|}, \tag{14}$$

where a target $y_i = 0$ gives an infinite term when $e_i \neq 0$ and an undefined term when $e_i = 0$ [31]. In Eq. (14), sMAPE is the symmetric variant with a per-case value of at most 2. The coefficient of determination compares the model with a constant prediction. Eq. (15) defines it.

$$R^2 = 1 - \frac{\sum_i (y_i - \hat{y}_i)^2}{\sum_i (y_i - \bar{y})^2}, \tag{15}$$

where $\bar{y}$ is the mean of the test targets. In Eq. (15), $R^2$ is negative when the model is worse than predicting $\bar{y}$ for every case [32].

### 2.3. Controlled scenario design

Each scenario isolates one mechanism. It fixes all inputs except one manipulated factor. Confusion-matrix scenarios lock the four counts. Score-based scenarios lock the label and score vectors, because one confusion matrix cannot determine an AUC or a calibration curve. Random inputs were drawn once with seed 20261102 (offsets +6, +7 and +9 for S6, S7 and S9) and saved as CSV files. These saved inputs are reused without regeneration. The model for each random input is specified in Appendix Table C1. Fold counts for S14 are provided in Appendix Table C2. A common template is used for all scenarios: purpose, locked inputs, manipulated factor, results, figure, interpretation and reporting recommendation. The 15 scenarios are arranged in four blocks in Table 2.

Table 2. Controlled scenarios S1–S15. The four blocks cover confusion matrices (A), scores and probabilities (B), regression (C), and inference and reporting (D). Fixed inputs are listed in Appendix Table A1.

| No | Block | Mechanism | Locked input | Manipulated factor |
|---|---|---|---|---|
| S1 | A | Accuracy paradox | two confusion matrices, prevalence 0.010 | classifier |
| S2 | A | Prevalence shift | TPR 0.800, TNR 0.900, T = 10,000 | prevalence |
| S3 | A | F1, MCC and positive class | TP 90, FN 10, FP 10; one 2 × 2 table | TN count; positive label |
| S4 | A | Multi-class averaging | one 3 × 3 confusion matrix | averaging rule |
| S5 | B | ROC and PR under low prevalence | 100 positive and 100 negative binormal scores | prevalence (0.5, 0.010) |
| S6 | B | Threshold and error cost | validation and test sets, 2,000 cases each | threshold rule |
| S7 | B | Discrimination and calibration | 5,000 logits and labels | logit scale |
| S8 | A | Kappa and marginal totals | two 2 × 2 tables, n = 100 | marginal distribution |
| S9 | C | Outliers in regression error | 100 targets, two error vectors | error distribution |
| S10 | C | Near-zero targets | 10 targets, error 1 | target magnitude |
| S11 | C | Negative $R^2$ | 21 targets, three predictors | reference prediction |
| S12 | D | Small test sets | accuracy 0.950 at three sizes | test size |
| S13 | D | Aggregation level | 10 participants, 700 segments | aggregation rule |
| S14 | D | Reverse calculation | eight reported triplets, five folds | reported values, rounding, aggregation |
| S15 | D | Software conventions | S1, S5 and S10 inputs, small label vectors | library option |

The mechanism and varied factor for each scenario are specified in Table 2. Possible effects are demonstrated with constructed inputs. Their frequency in practice is not estimated.

Thresholds for S6 were selected on a validation set. They were then applied without change to a separate test set of equal size. Test data were excluded from threshold selection. For the low-prevalence set in S5, the same 100 positive scores were retained. Each negative score was repeated 99 times. Class-conditional score distributions were thus identical to those in the balanced set. Labels in S7 were drawn as $y_i \sim \mathrm{Bernoulli}(\mathrm{sigmoid}(z_i))$. Thus, $\mathrm{sigmoid}(z_i)$ is the true event probability. The variant with $a = 1$ is calibrated by construction.

### 2.4. Uncertainty and model comparison

A proportion such as accuracy, sensitivity or specificity is a binomial estimate when cases are independent. The Wilson score interval is defined in Eq. (16) [33].

$$\frac{\hat{p} + \frac{z^2}{2n}}{1 + \frac{z^2}{n}} \pm \frac{z}{1 + \frac{z^2}{n}} \sqrt{\frac{\hat{p}(1 - \hat{p})}{n} + \frac{z^2}{4n^2}}, \tag{16}$$

where $\hat{p} = k/n$ is the observed proportion of $k$ successes in $n$ cases and $z = 1.96$ for a 95% interval. In Eq. (16), the interval stays between 0 and 1. The exact Clopper–Pearson interval uses beta quantiles [34]. Both intervals have better coverage than the normal approximation [35]. Repeated segments of one participant are not independent [40]. For S13, participants were therefore resampled with replacement (10,000 resamples, seed 20261115) and percentile intervals were computed; the participant-majority proportion received an exact Clopper–Pearson interval. Table 3 lists the methods for other quantities, their assumptions and whether each was applied here.

Table 3. Interval and test methods by quantity. Independent cases are assumed unless a cluster method is named. Status: whether the method was applied to a scenario in this study.

| Quantity | Method | Assumption | Status |
| --- | --- | --- | --- |
| Proportion (accuracy, sensitivity, specificity, precision) | Wilson score interval; exact Clopper–Pearson interval | independent cases; fixed denominator | applied (S12, S13) |
| Paired decisions of two classifiers on one test set | McNemar's test | independent cases; same test set | recommended; not applied |
| Two correlated ROC-AUCs | DeLong's method | independent cases; same test set | recommended; not applied |
| MCC, F1, AP, Brier, ECE and other functions of the data | percentile bootstrap | resampling unit is the independent unit | recommended; not applied |
| Repeated measurements of one participant | cluster bootstrap over participants | participants are independent | applied (S13) |
| Scores averaged over cross-validation folds | no exact interval; fold results reported | folds share training data | illustrated (S14) |

Table 3 links each quantity to an interval or test and to the assumption behind it. Methods marked as recommended were not applied to any scenario in this study.

### 2.5. Consistency assessment framework

Reported values can be checked against each other. Eq. (1) implies that accuracy lies between sensitivity and specificity. When TPR ≠ TNR, it also implies the prevalence in Eq. (17).

$$\pi = \frac{\text{Acc} - \text{TNR}}{\text{TPR} - \text{TNR}}, \tag{17}$$

where all three values come from one test set, one threshold and one aggregation level. In Eq. (17), rounding enters through the denominator. Each reported value was therefore treated as an interval of half a unit of its last reported decimal, and the reporting precision of every case is kept in Table 14. The implied prevalence was evaluated over that interval box rather than at one point.

When the class counts $P$ and $N$ are known, a stronger check enumerates all integer pairs $(TP, TN)$ and keeps those that reproduce every reported value within its rounding interval. An empty set means that the values cannot belong to one confusion matrix. The same tests were run with mlscorecheck 1.0.3 [45]. Its fold-wise model treats scores as means over folds (mean of scores), and its single-test-set model treats them as computed on one pooled confusion matrix [44]. Table 4 defines the four outcome classes used to report a check.

Table 4. Outcome classes for a consistency check. A check is made under a stated model of the experiment: one test set or a mean over folds. Examples refer to the cases of S14 (Table 14).

| Outcome | Condition | Example in S14 |
|---|---|---|
| Consistent | at least one confusion matrix reproduces the values exactly | case A: 1 exact and 9 rounding-compatible (TP, TN) pairs |
| Consistent within rounding | a confusion matrix reproduces the values only within half a unit of the last decimal | case E: 0 exact and 1 rounding-compatible pair |
| Inconsistent | no confusion matrix reproduces the values under the stated model | case B (accuracy outside the interval between TPR and TNR); case D |
| Not assessable | the information or the model does not allow a check | case C2 (TPR close to TNR); fold means tested as one pooled test set |

Table 4 separates a failed check from a check that could not be made. A report classed as inconsistent is not thereby shown to contain an error in the underlying experiment.

### 2.6. Implementation and verification

All metrics were implemented from their formulas in Python. The implementation returns not-a-number whenever a denominator is zero instead of applying a convention. Table 1 and Table 15 classify each such case as undefined (0/0) or infinite. The scenario builder wrote the locked inputs and all results to one results file. A separate script recomputed the reported values with scikit-learn 1.9.1 [46], statsmodels 0.15.0 [47], SciPy 1.17.1 [48] and mlscorecheck 1.0.3 [45]. Each check was assigned to one of three kinds. An independent library check recomputed a value with a function of another library, either exactly or within a Monte Carlo tolerance when random resampling was involved. An independent re-implementation coded the same formula separately, for example the ECE weighting on library bin statistics. An internal check re-derived a value from the input file with the same arithmetic and is not counted as independent. Monte Carlo checks used 20,000 and 2,000 random permutations for the expected AP of the two S5 sets (seed 7) and 10,000 SciPy bootstrap resamples for the S13 intervals (seed 11). Appendix Table B1 lists every check group with its reference implementation, resampling settings and tolerance. The scenario specification and all tables in this paper were generated from the same results file, so no value was typed by hand.

### 2.7. Scope and article type

This study is a methodological tutorial. Its evidence consists of constructed scenarios with released inputs and verified arithmetic. Two empirical extensions were planned but not carried out. These were a literature audit of reporting practice and a demonstration with public real-data tasks. Both are specified in Appendix D so that they can be run and reported separately.

## 3. Results

The scenarios were run as specified in Section 2.3, and the values were recomputed as described in Section 2.6. All 205 checks passed (Appendix Table B1). Of these, 178 were exact independent library checks, 6 were Monte Carlo library checks, 18 were independent re-implementations and 3 were internal re-derivations. Values are proportions on the 0–1 scale unless a count or a cost is stated.

### 3.1. Confusion-matrix scenarios

The accuracy paradox appeared as constructed in S1 (Table 5).

Table 5. Scenario S1, accuracy paradox. T = 10,000 cases with 100 positive cases (prevalence 0.010); the positive class is the minority class. Counts are TP / FN / FP / TN.

| Rule | Counts | Acc | BA | TPR | TNR | PPV | F1 | MCC |
|---|---|---|---|---|---|---|---|---|
| Always negative | 0 / 100 / 0 / 9,900 | 0.990 | 0.500 | 0.000 | 1.000 | undefined | 0.000 | undefined |
| Screening model | 80 / 20 / 495 / 9,405 | 0.949 | 0.875 | 0.800 | 0.950 | 0.139 | 0.237 | 0.321 |

Table 5 shows that accuracy ranked the always-negative rule above the screening model (0.990 versus 0.949). Balanced accuracy reversed that order (0.500 versus 0.875), and the always-negative rule had undefined precision and MCC.

The screening model detected 80 of 100 positive cases. Its precision was only 0.139, because 495 false positives arose from 9,900 negative cases. Accuracy therefore rewarded the rule that ignored the positive class, whereas the class-wise rates exposed it.

Prevalence changed every gold-coded metric of Figure 2 in S2, while sensitivity and specificity stayed at 0.800 and 0.900 (Figure 3).

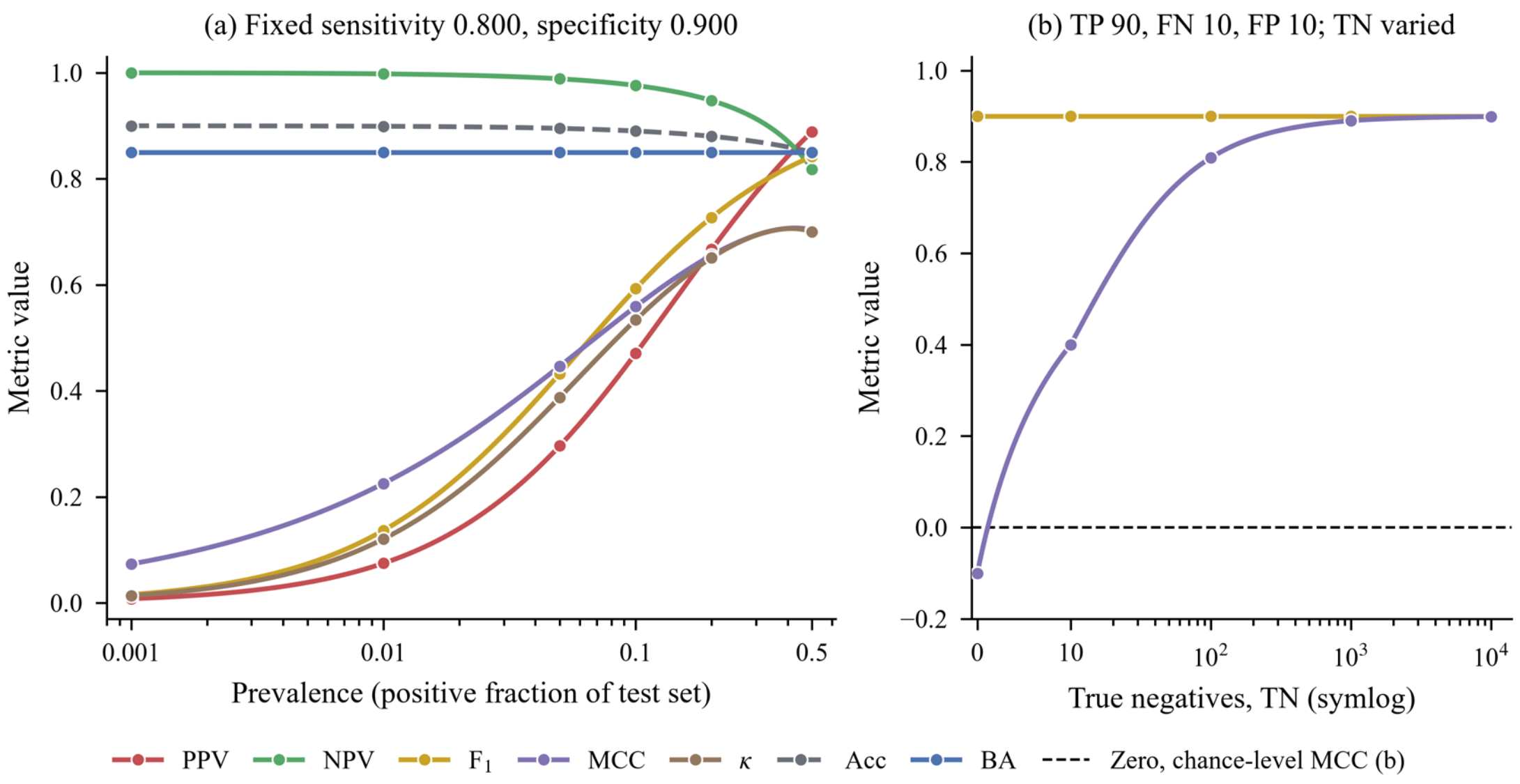

Fig. 3. Dependence of confusion-matrix metrics on prevalence and on the true-negative count (scenarios S2 and S3a). (a) PPV, NPV, F1, MCC, κ, accuracy (Acc, dashed) and balanced accuracy (BA) against test-set prevalence (log axis) at fixed sensitivity 0.800 and specificity 0.900; lines are analytic curves and markers are the six locked prevalences 0.001, 0.01, 0.05, 0.1, 0.2 and 0.5 (T = 10,000 each); (b) F1 and MCC against the number of true negatives (TN = 0, 10, 100, 1,000 and 10,000; symmetric-log axis, linear between 0 and 10) with TP 90, FN 10 and FP 10; lines are analytic curves and markers are the locked tables, and the black dashed line marks zero (chance-level MCC). In S3a κ equals MCC to three decimals and is therefore not drawn in (b). Each metric keeps one colour in all figures (Acc grey dashed, BA blue, PPV red, NPV green, F1 gold, MCC purple, κ brown).

Figure 3 shows that precision fell from 0.889 at a prevalence of 0.5 to 0.008 at 0.001, whereas balanced accuracy stayed at 0.850. These curves follow from the fixed class-wise rates and do not describe a classifier whose behaviour changes with the class mix.

Accuracy rose from 0.850 to 0.900 as prevalence fell, because the specificity of the larger negative class exceeded the sensitivity. The F1 score fell from 0.842 to 0.016, MCC from 0.704 to 0.074 and kappa from 0.700 to 0.014. Such values can be compared across test sets only when the target population and the evaluation purpose are stated. Differences in these values cannot alone be attributed to differences between models.

True negatives were excluded from F1 in S3 but were included in MCC and kappa (Table 6).

Table 6. Scenario S3: F1, MCC and the choice of positive class. (a) TN is varied at fixed TP 90, FN 10 and FP 10. (b) The same 2 × 2 table is evaluated with each class designated as positive. Counts are listed as TP / FN / FP / TN.

| Setting | Counts | Acc | F1 | MCC | κ |
|---|---|---|---|---|---|
| (a) TN = 0 | 90 / 10 / 10 / 0 | 0.818 | 0.900 | −0.100 | −0.100 |

| Setting | Counts | Acc | F1 | MCC | κ |
|---|---|---|---|---|---|
| (a) TN = 10 | 90 / 10 / 10 / 10 | 0.833 | 0.900 | 0.400 | 0.400 |
| (a) TN = 100 | 90 / 10 / 10 / 100 | 0.905 | 0.900 | 0.809 | 0.809 |
| (a) TN = 1,000 | 90 / 10 / 10 / 1,000 | 0.982 | 0.900 | 0.890 | 0.890 |
| (a) TN = 10,000 | 90 / 10 / 10 / 10,000 | 0.998 | 0.900 | 0.899 | 0.899 |
| (b) positive = class 1 | 40 / 10 / 20 / 930 | 0.970 | 0.727 | 0.715 | 0.712 |
| (b) positive = class 0 | 930 / 20 / 10 / 40 | 0.970 | 0.984 | 0.715 | 0.712 |

In Table 6, F1 remained at 0.900 as TN increased from 0 to 10,000. MCC increased from −0.100 to 0.899. After the class labels were reversed, F1 changed from 0.727 to 0.984. MCC remained at 0.715. The positive class must therefore be stated with every F1 value.

Several F1 summaries were calculated from the same three-class confusion matrix in S4 (Table 7).

Table 7. Scenario S4: alternative averages for the same three-class confusion matrix. Class supports are A 800, B 150 and C 50. Gorodkin's R_K is used for MCC. BA is the macro average of recall.

| Acc | Micro F1 | Macro F1 | Weighted F1 | BA | MCC | κ | F1 A | F1 B | F1 C |
|---|---|---|---|---|---|---|---|---|---|
| 0.890 | 0.890 | 0.707 | 0.888 | 0.694 | 0.665 | 0.664 | 0.944 | 0.733 | 0.444 |

In Table 7, micro F1 and accuracy were both 0.890. Macro F1 was 0.707. This gap was due to minority class C, with an F1 of 0.444. All summaries were obtained from the same classifier. The weighted average (0.888) stayed close to the micro average, because class A held 800 of the 1,000 cases. Balanced accuracy, the macro average of recall, was 0.694. An F1 value without its averaging rule is therefore not interpretable in multi-class work.

Identical observed agreement gave different kappa values in S8 (Table 8).

Table 8. Scenario S8, kappa and marginal totals (n = 100 in each table). Counts are TP / FN / FP / TN; p_e is the chance agreement of Eq. (6).

| Table | Counts | Observed agreement | p_e | κ | MCC |
|---|---|---|---|---|---|
| Balanced marginals | 40 / 6 / 9 / 45 | 0.850 | 0.501 | 0.700 | 0.701 |
| Skewed marginals | 80 / 5 / 10 / 5 | 0.850 | 0.780 | 0.318 | 0.327 |

Table 8 shows an observed agreement of 0.850 in both tables, with kappa values of 0.700 and 0.318. The lower value reflects the higher chance agreement implied by skewed marginal totals and does not indicate worse raters.

### 3.2. Score and probability scenarios

ROC-AUC did not change with prevalence in S5, whereas average precision did (Figure 4).

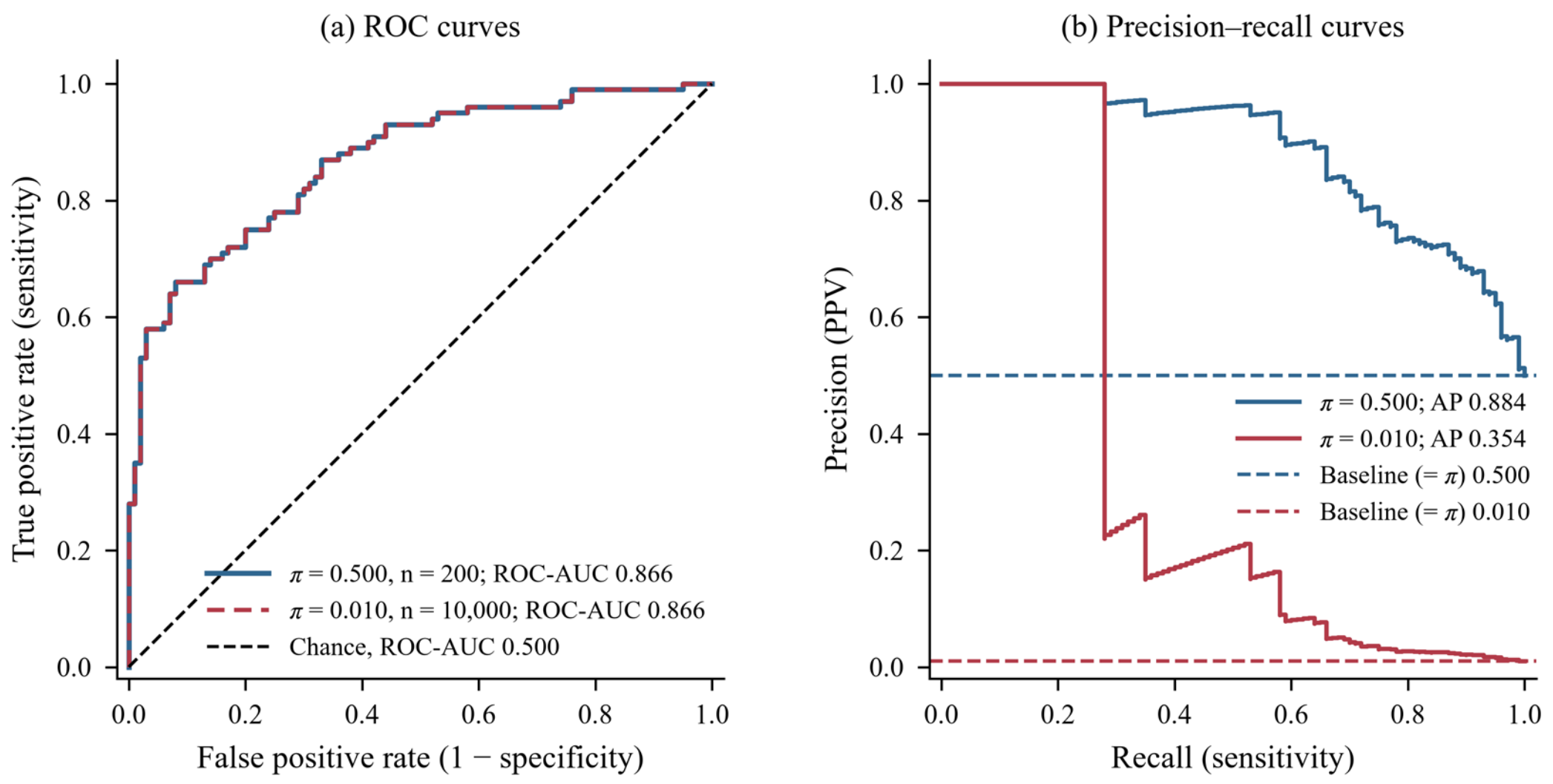

Fig. 4. ROC and precision–recall curves at two prevalences with identical class-conditional score distributions (scenario S5). (a) ROC curves of the balanced set (π = 0.500, n = 200; solid blue) and the low-prevalence set (π = 0.010, n = 10,000; dashed red), which coincide; the black dashed diagonal is the chance line; (b) step precision–recall curves of the same two sets, each with its population reference line (precision equal to the prevalence π) dashed in the matching colour; the expected AP of a random ranking in these finite sets is slightly higher (Section 3.2). Legends give ROC-AUC and average precision (AP, step-wise scikit-learn estimator). Positive scores follow N(1.5, 1) and negative scores N(0, 1), 100 of each; the low-prevalence set repeats each negative 99 times.

Figure 4 shows a ROC-AUC of 0.866 in both sets and an AP that fell from 0.884 to 0.354. The AP of the low-prevalence set lay far above its population reference of 0.010, so the ranking remained informative even though precision was low. The expected AP of a random ranking was 0.512 in the balanced set and 0.011 in the low-prevalence set, slightly above the respective prevalence.

Five operating points were evaluated from the same score vector in S6 (Table 9).

Table 9. Scenario S6: comparison of threshold rules. Thresholds were chosen on the validation set and applied to the test set (2,000 cases each, prevalence 0.10). All metrics are test-set values. Cost per 1,000 test cases uses a false-negative cost 20 times the false-positive cost.

| Rule (chosen on validation data) | Threshold | TPR | TNR | PPV | F1 | Cost per 1,000 |
|---|---|---|---|---|---|---|
| Default 0.5 | 0.500 | 0.820 | 0.940 | 0.603 | 0.695 | 414.0 |
| Youden | 0.415 | 0.890 | 0.902 | 0.503 | 0.643 | 308.0 |
| Maximum F1 | 0.568 | 0.780 | 0.963 | 0.700 | 0.738 | 473.5 |
| TPR ≥ 0.95 | 0.280 | 0.950 | 0.807 | 0.353 | 0.515 | 274.0 |
| Minimum cost | 0.353 | 0.920 | 0.867 | 0.435 | 0.591 | 279.5 |

Table 9 shows test-set sensitivities from 0.780 to 0.950 for the same scores. The threshold that minimized cost on validation data gave a test cost of 279.5 per 1,000 cases, slightly above the 274.0 of the sensitivity-constrained rule; validation optimality is therefore not test optimality. The validation constraint of a sensitivity of at least 0.95 was met on the test set (0.950) at a precision of 0.353. The F1-optimal threshold of 0.568 gave the highest F1 (0.738) but the highest cost (473.5). A threshold is therefore a design choice that must be reported with its rule, its data split and, where used, its cost ratio.

Ranking and decisions were identical across the three probability variants of S7, whereas probability quality differed (Table 10).

Table 10. Scenario S7: discrimination and calibration. Predicted probabilities are defined as sigmoid(a · z). The same fixed logit vector z is used for all 5,000 cases. The variant a = 1 is calibrated by construction. ECE uses 10, 15 or 20 equal-width bins.

| a | ROC-AUC | Acc at 0.5 | Brier | Log loss | ECE 10 | ECE 15 | ECE 20 | Slope | Intercept |
|---|---|---|---|---|---|---|---|---|---|
| 0.5 | 0.825 | 0.769 | 0.170 | 0.519 | 0.106 | 0.106 | 0.106 | 2.08 | −0.272 |
| 1.0 | 0.825 | 0.769 | 0.157 | 0.477 | 0.013 | 0.017 | 0.019 | 1.04 | 0.020 |
| 2.0 | 0.825 | 0.769 | 0.168 | 0.547 | 0.094 | 0.094 | 0.095 | 0.52 | 0.479 |

All variants in Table 10 had a ROC-AUC of 0.825. Accuracy was also identical at 0.769 for a threshold of 0.5. Differences were captured by the Brier score, log loss, ECE and calibration slope. Probability quality therefore cannot be assessed from a discrimination metric alone.

The calibrated variant had a calibration slope of 1.04 and an ECE of 0.013 with 10 bins. Its ECE rose to 0.019 with 20 bins. This change was caused solely by the number of bins. The reliability diagrams are presented in Figure 5.

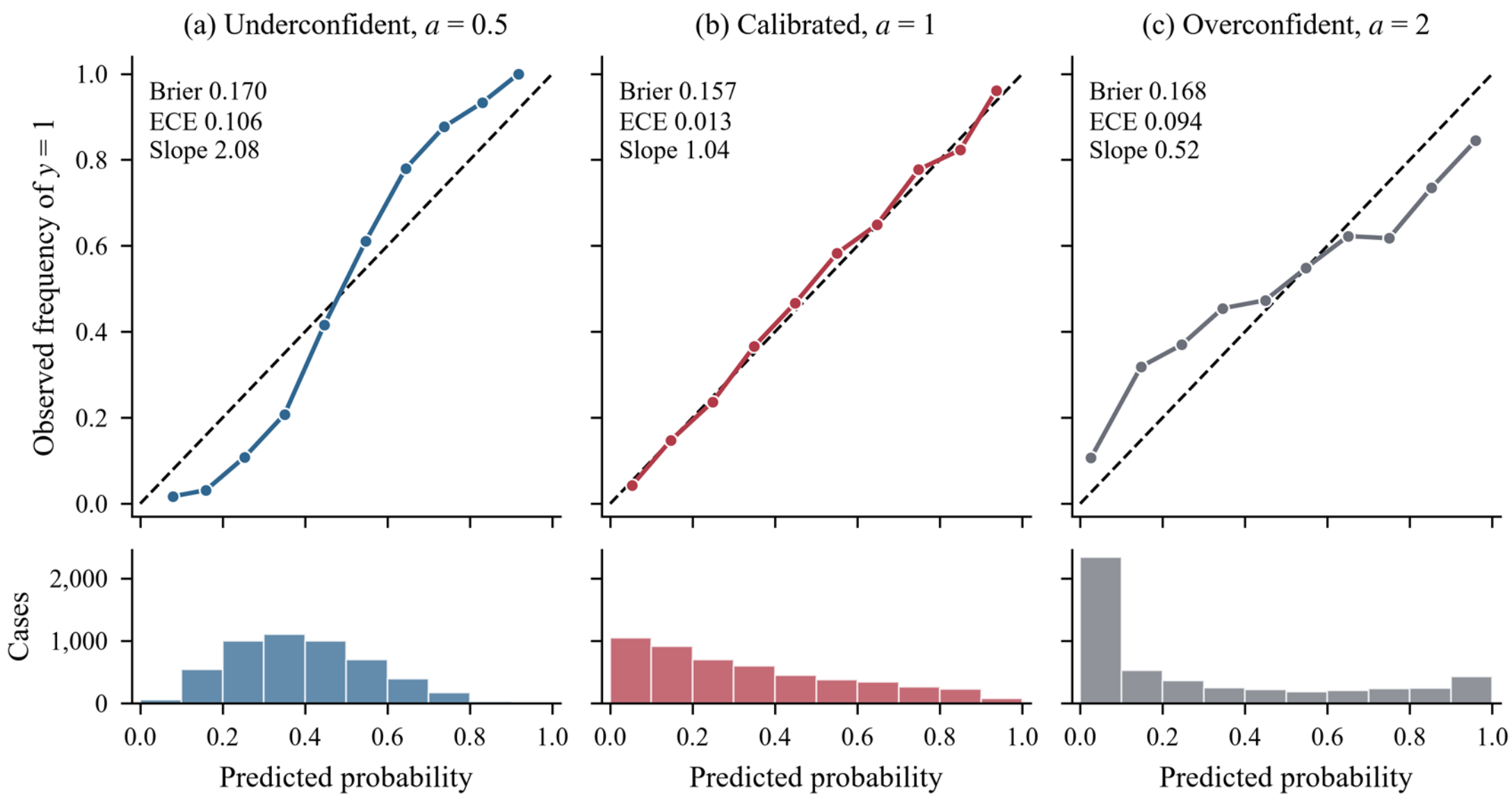

Fig. 5. Reliability diagrams for three probability variants with the same score order (scenario S7, n = 5,000). Probabilities are p = sigmoid(a·z) for logit scale (a) a = 0.5 (underconfident, blue), (b) a = 1 (calibrated, red) and (c) a = 2 (overconfident, grey). Upper panels show the observed frequency of y = 1 against the mean predicted probability in 10 equal-width bins, with the identity line dashed black; lower strips show the number of cases per bin. Each panel prints the Brier score, the ECE (10 equal-width bins) and the calibration slope (logistic regression of y on logit(p)). The highest bin in (a) contains 1 case.

Figure 5 shows the underconfident variant below the identity line at low probabilities and above it at high probabilities, with the reverse pattern for the overconfident variant. Reliability diagrams describe calibration on this test set and do not guarantee calibration under a different class mix.

### 3.3. Regression scenarios

MAE and RMSE ranked the two S9 models in opposite order (Table 11).

Table 11. Regression scenarios S9–S11. Model A has an absolute error of 2 for all 100 cases; model B has 0.5 for 95 cases and 30 for 5 cases. (a) MAPE over the nine non-zero targets; the percentage error of the zero target is infinite. r = Pearson correlation, undefined for a constant prediction.

| Scenario | Predictor | MAE | RMSE | MedAE | MAPE | $R^2$ | r |
|---|---|---|---|---|---|---|---|
| S9 | Model A | 2.000 | 2.000 | 2.000 | 0.045 | 0.987 | 0.99 |
| S9 | Model B | 1.975 | 6.726 | 0.500 | 0.049 | 0.849 | 0.94 |
| S10 | Target + 1 | 1.000 | 1.000 | 1.000 | 0.432 (a) | 1.000 | 1.00 |
| S11 | Target + 15 | 15.000 | 15.000 | 15.000 | 0.305 | −5.136 | 1.00 |
| S11 | Constant 45 (training mean) | 6.429 | 7.853 | 5.000 | 0.121 | −0.682 | undefined |
| S11 | Constant 50 (test mean) | 5.238 | 6.055 | 5.000 | 0.107 | 0.000 | undefined |

Table 11 shows a slightly lower MAE for model B (1.975 versus 2.000) and a much higher RMSE (6.726 versus 2.000). Model B's MedAE of 0.500 hid its five large errors, so a single absolute-error summary did not describe the error distribution.

A constant absolute error of 1 produced very different percentage errors in S10. The absolute percentage error was 2.000 at a target of 0.5 and 0.005 at a target of 200. At a target of 0 it was infinite, so the reported MAPE of 0.432 required the zero target to be excluded.

A perfectly correlated but biased predictor in S11 had a Pearson correlation of 1.00 and an $R^2$ of −5.136. Predicting the training mean of 45 gave −0.682, and predicting the test mean gave exactly 0.000. Figure 6 shows the three regression scenarios together.

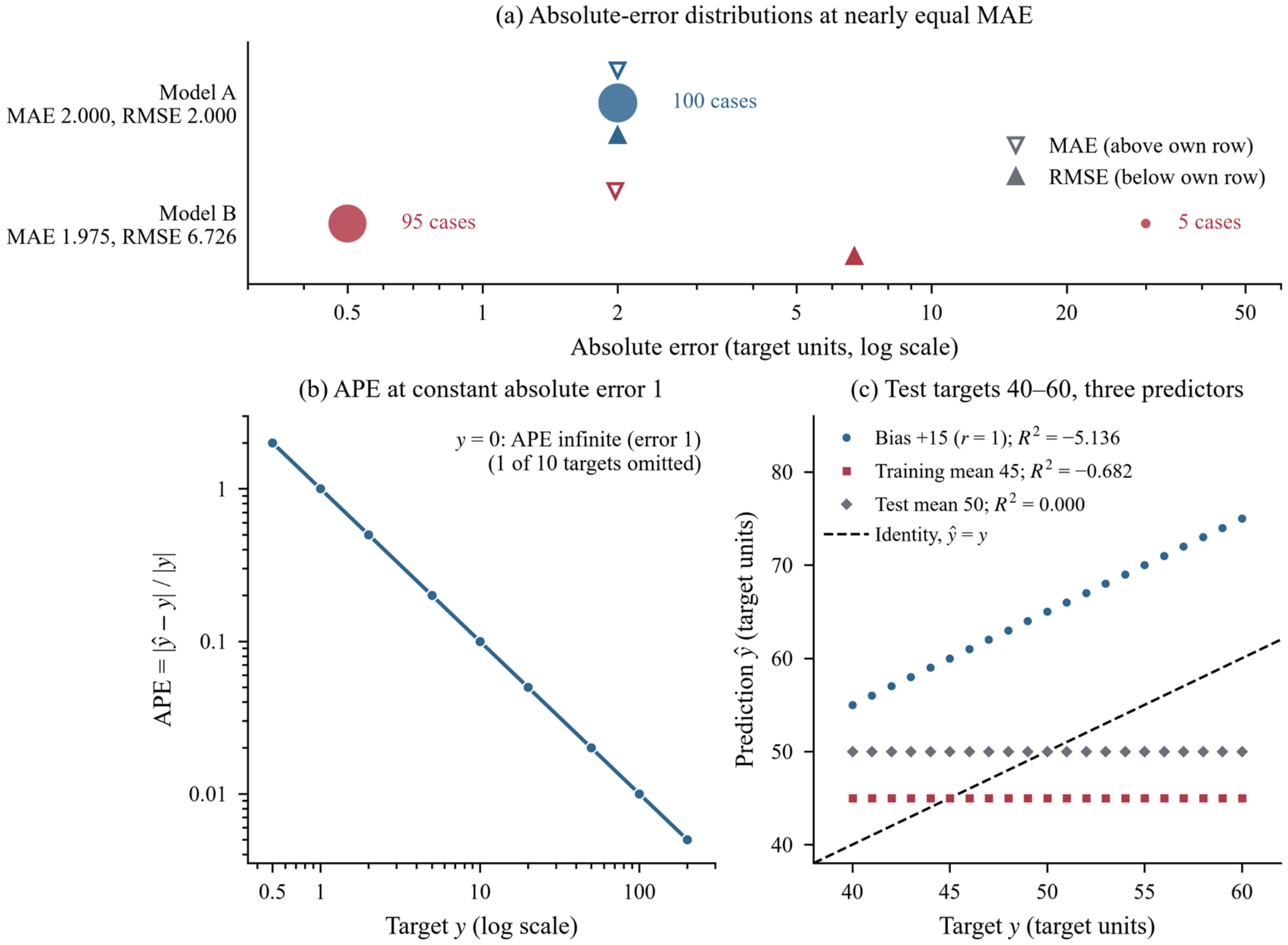

Fig. 6. Regression error metrics under outliers, near-zero targets and biased prediction (scenarios S9–S11). (a) Absolute errors of model A (blue; 100 cases with error 2) and model B (red; 95 cases with error 0.5 and 5 cases with error 30) on a log axis; marker area scales with the number of cases, open downward triangles above each row mark MAE and filled upward triangles below each row mark RMSE, and the row labels give both values; (b) absolute percentage error, APE = |ŷ − y| / |y| on the 0–1 scale, against target y on log–log axes for a constant absolute error of 1 (S10); APE is infinite at y = 0 (error 1), so that target is omitted; (c) predictions against test targets 40–60 for a predictor with bias +15 (blue circles), the constant training mean 45 (red squares) and the constant test mean 50 (grey diamonds), with the identity line dashed black and $R^2$ in the legend.

Figure 6 shows that the error distribution, the target scale and the reference prediction each changed the reading of a regression metric. Its panels use constructed vectors and do not rank regression metrics in general.

### 3.4. Inference and reporting scenarios

The same accuracy carried very different uncertainty in S12 (Table 12).

Table 12. Scenario S12, 95% intervals for an accuracy of 0.950 at three test sizes. CP = exact Clopper–Pearson interval.

| Correct / n | Acc | Wilson 95% CI | CP 95% CI | Wilson half-width |
|---|---|---|---|---|
| 38 / 40 | 0.950 | 0.835–0.986 | 0.831–0.994 | 0.076 |
| 380 / 400 | 0.950 | 0.924–0.967 | 0.924–0.969 | 0.022 |

| Correct / n | Acc | Wilson 95% CI | CP 95% CI | Wilson half-width |
|---|---|---|---|---|
| 3,800 / 4,000 | 0.950 | 0.943–0.956 | 0.943–0.957 | 0.007 |

Table 12 shows a 95% Wilson interval of 0.835–0.986 for 38 of 40 correct cases and 0.943–0.956 for 3,800 of 4,000. These intervals assume independent cases and are too narrow for repeated measurements of the same participant.

A Wilson half-width of at most 0.02 at an accuracy of 0.95 required 480 test cases, and a half-width of 0.01 required 1,840. Two models whose accuracies differ by a few hundredths cannot be separated on a test set of 40 cases.

The aggregation rule changed accuracy in S13, and the uncertainty depended on the resampling unit (Table 13).

Table 13. Scenario S13, accuracy at three aggregation levels (10 participants, 700 segments). Cluster bootstrap: participants resampled with replacement, 10,000 resamples, seed 20261115, percentile intervals. The naive Wilson interval treats segments as independent and is shown for contrast only.

| Summary | Estimate | 95% interval | Interval method |
|---|---|---|---|
| Segment-level pooled accuracy | 0.857 | 0.677–0.916 | cluster bootstrap |
| Segment-level pooled accuracy | 0.857 | 0.829–0.881 | naive Wilson (segments as independent) |
| Mean per-participant accuracy | 0.691 | 0.556–0.825 | cluster bootstrap |
| Participant-majority accuracy | 0.700 (7 / 10) | 0.348–0.933 | exact Clopper–Pearson |

Table 13 shows a pooled segment-level accuracy of 0.857, a mean per-participant accuracy of 0.691 and a participant-majority accuracy of 0.700. The cluster-bootstrap interval of the pooled accuracy (0.677–0.916) was much wider than the naive Wilson interval (0.829–0.881), which wrongly treats segments as independent.

The participant-majority accuracy of 7 of 10 had an exact interval of 0.348–0.933. Ten participants therefore cannot distinguish this result from chance-level majority decisions. Figure 7 shows both inference scenarios.

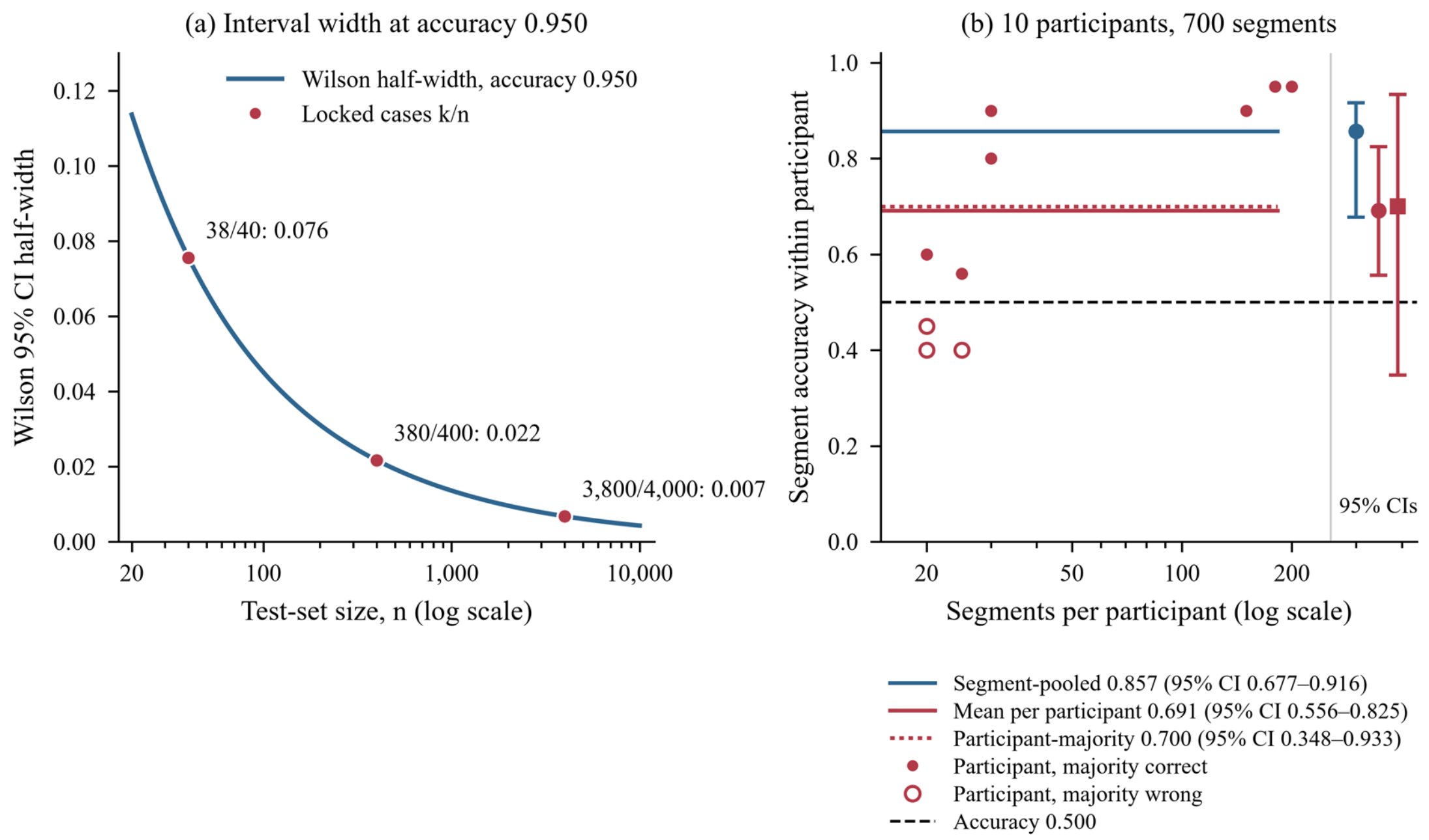

Fig. 7. Uncertainty of accuracy and dependence on the aggregation level (scenarios S12 and S13). (a) Half-width of the 95% Wilson interval against test-set size n (log axis, 20–10,000) at accuracy 0.950 (blue line), with the three locked cases 38/40, 380/400 and 3,800/4,000 (red markers, half-widths printed); (b) segment accuracy within each participant against the participant's number of segments (log axis; 10 participants, 700 segments); filled red markers are participants whose segment majority is correct and open markers those whose majority is wrong. Horizontal lines give the segment-pooled accuracy (blue), the mean per-participant accuracy (red, solid) and the participant-majority accuracy (fraction of participants with a correct majority; red, dotted); the black dashed line marks 0.5. Error bars to the right of the grey line give 95% intervals: cluster bootstrap over participants for the pooled and mean accuracies and exact Clopper–Pearson for the participant-majority accuracy (Table 13).

Figure 7 shows that interval width shrinks with the square root of the test size and that three participants with many segments dominated the pooled estimate. Its intervals assume that the 10 constructed participants are independent and say nothing about other cohorts.

Reverse calculation separated possible from impossible reports in S14 (Table 14).

Table 14. Scenario S14, reverse calculation of reported triplets. Values are shown with their reported precision; the tolerance is half a unit of the last reported decimal and applies to all three values. A4 repeats A at four decimals. Pairs: (TP, TN) pairs compatible with all three values within the stated tolerance. Exact pairs are given in parentheses. Implied π: the range obtained from Eq. (17) within the tolerance box.

| Case | P | N | Acc | TPR | TNR | Tolerance | Pairs (exact) | Implied π | Outcome |
|---|---|---|---|---|---|---|---|---|---|
| A | 100 | 900 | 0.89 | 0.80 | 0.90 | ±0.005 | 9 (1) | 0.000–0.200 | consistent |
| A4 | 100 | 900 | 0.8900 | 0.8000 | 0.9000 | ±0.00005 | 1 (1) | 0.099–0.101 | consistent |
| B | 100 | 900 | 0.95 | 0.90 | 0.92 | ±0.005 | 0 (0) | none | inconsistent |
| C1 | unknown | unknown | 0.83 | 0.95 | 0.80 | ±0.005 | not computed | 0.133–0.267 | counts not assessable; prevalence bounded |
| C2 | unknown | unknown | 0.90 | 0.91 | 0.90 | ±0.005 | not computed | 0.000–1.000 | counts not assessable; prevalence not assessable |
| D | 50 | 150 | 0.9133 | 0.8400 | 0.9400 | ±0.00005 | 0 (0) | 0.266–0.268 | inconsistent |
| D | 50 | 150 | 0.9150 | 0.8400 | 0.9400 | ±0.00005 | 1 (1) | 0.249–0.251 | consistent |
| E | 30 | 170 | 0.890 | 0.833 | 0.900 | ±0.0005 | 1 (0) | 0.134–0.164 | consistent within rounding |

Case B in Table 14 was impossible. Its accuracy was outside the interval bounded by sensitivity and specificity. For case D, the reported accuracy was 0.9133. This was 0.0017 below the feasible value of 0.9150. Such a discrepancy can be detected by integer enumeration. It may be missed in a visual review of three plausible values.

The consistency result depended on the number of reported decimals. Nine (TP, TN) pairs were compatible with case A at two decimals. Only one pair remained for the same values at four decimals in case A4. Case E was consistent only after allowance for numerical precision. Its sensitivity of 0.833 was compatible with 25 of 30 cases (0.8333) at three decimals. The same value written as 0.8330 would have been inconsistent.

The implied prevalence was informative only when sensitivity and specificity were far apart. It lay between 0.133 and 0.267 for case C1 and covered the whole interval from 0 to 1 for case C2. Even the consistent case A, reported to two decimals, fixed the prevalence only to the range from 0 to 0.200.

Fold averages broke the pooled identity of Eq. (1). The mean fold scores implied a prevalence of 0.2126, whereas the pooled prevalence was 0.2000. mlscorecheck found the mean fold scores consistent under its fold-wise model. It flagged them as inconsistent when they were tested as one pooled test set. A check under the wrong aggregation model therefore produced a false alarm, which Table 4 classes as not assessable.

Software options changed reported values for identical predictions in S15 (Table 15).

Table 15. Scenario S15, software conventions in scikit-learn 1.9.1 for identical predictions. Status: the mathematical value of the quantity (finite, infinite or undefined). Library values are listed in the order of the options; a value that replaces an infinite or undefined quantity is assigned by the library.

| Quantity | Status | Options | Library values |
|---|---|---|---|
| Precision with no positive predictions | undefined (0/0) | zero_division = warn / 0 / 1 / nan | 0.000 / 0.000 / 1.000 / undefined |
| Macro F1 with class 2 absent | per-class F1 of class 2 undefined (0/0) | labels inferred / labels fixed to {0, 1, 2} | 0.667 / 0.444 |
| MCC of the always-negative rule (S1) | undefined (0/0) | formula / library | undefined / 0.000 |
| PR summary, S5 low-prevalence set, continuous scores | finite; estimator choice | AP (step) / trapezoidal area | 0.354 / 0.353 |
| PR summary, same set, 7 distinct scores | finite; estimator choice | AP (step) / trapezoidal area | 0.230 / 0.412 |
| MAPE with one zero target (S10) | infinite (error 1, target 0) | eps convention / zero target excluded | $4.5 \times 10^{14}$ / 0.432 |

Table 15 shows that an undefined precision was reported as 0, 1 or not a number depending on one option. An infinite percentage error at a zero target became $4.5 \times 10^{14}$ under the eps convention of scikit-learn. These values are assigned by library conventions; they are neither errors of the library nor mathematical values of the metric.

The precision–recall summary depended on the estimator once scores were coarse. With continuous scores, AP and the trapezoidal area were 0.354 and 0.353. After rounding to 7 distinct values, they were 0.230 and 0.412. Linear interpolation therefore overstated the precision–recall summary in this example.

## 4. Discussion

### 4.1. Main findings

The scenarios showed that one set of predictions supports several defensible numbers. The principal findings are summarized below.

- Accuracy, precision, NPV, F1, MCC and kappa changed with prevalence at fixed sensitivity and specificity, whereas balanced accuracy, informedness and the likelihood ratios did not.
- F1 did not use true negatives and changed when the class labels were swapped, whereas MCC used all four counts and was symmetric.

- ROC-AUC and AP were unchanged by any strictly increasing transformation of the scores, because both depend only on the ranking. AP changed with prevalence, and the Brier score, log loss and ECE changed with the probability transformation.
- Thresholds chosen on validation data did not carry exact guarantees to the test set, and the F1-optimal threshold differed from the cost-optimal threshold.
- MAE, RMSE and MedAE ranked models differently when errors had heavy tails, and the percentage error was infinite at a zero target.
- The resampling unit, the reporting precision, the known class counts and the aggregation model decided how wide an interval was and whether a consistency check could be made at all.

All values were recomputed, but they remain properties of constructed inputs. The findings describe mechanisms and do not measure how often these mechanisms affect published results.

### 4.2. Interpretation and scope of the synthetic evidence

Each scenario was built to isolate one mechanism. The effect sizes were determined by the selected inputs. They should not be treated as typical values for published studies. A prevalence of 0.010 was selected for S1 and a bias of 15 units for S11. These choices made the mechanisms easy to observe.

The score-based scenarios depended on specified distributions. Binormal scores were used in S5. Negative scores were repeated in the low-prevalence set to preserve the class-conditional distributions. In real data, these distributions rarely remain fixed after a prevalence change. Case mix and measurement conditions may also vary [27]. A metric unchanged by class mix is therefore not guaranteed to transfer between datasets.

A monotone transformation of a well-calibrated logit was used in the probability scenarios. Real miscalibration may be non-monotone or class-dependent. It may also arise from dataset shift [24]. ECE in S7 depended on the equal-width bin rule. This dependence was demonstrated directly in the scenario.

Random inputs were drawn once with fixed seeds. Different values would be obtained in S5–S7 and S9 with other seeds. The direction of each effect follows from the formulas in Section 2.2. Variation across seeds was not quantified. Ten constructed participants were used in S13. The arithmetic of aggregation and bootstrap resamples was demonstrated. This scenario does not represent a validation design.

### 4.3. Metric choice follows the application goal

The sensitivity of each metric to specific changes is demonstrated by the scenarios. Metric choice must be guided by the application. For screening, missed positive cases may be the main concern. Sensitivity should then be reported at a stated threshold. For confirmation, precision should be assessed at the prevalence of the target population. Risk probabilities require calibration assessment and a proper scoring rule. These cannot be supplied by a ranking metric. Common application goals are linked to primary and complementary metrics in Table 16.

Table 16. Metric choice by application goal. The main application question is addressed by the primary metric. Additional information is provided by complementary metrics. The threshold, target-population prevalence and evaluation unit must be stated for every entry.

| Application goal | Primary metric | Complementary metrics | Scenario |
|---|---|---|---|
| Screening, missed positives costly | sensitivity at a pre-specified threshold | specificity, PPV at the target prevalence, cost at the threshold | S2, S6 |
| Confirmation, false alarms costly | PPV at the target prevalence | sensitivity, FDR, the prevalence used | S2 |
| Balanced overall decisions under imbalance | MCC or balanced accuracy | sensitivity, specificity, confusion matrix | S1, S3 |
| Retrieval of rare positives | average precision with its baseline | precision at a fixed recall or at k, ROC-AUC | S5, S15 |
| Ranking without a fixed threshold | ROC-AUC | ROC curve, average precision when positives are rare | S5 |
| Probabilities used for decisions or risk communication | log loss or Brier score | calibration slope and intercept, reliability diagram, ECE with its bins | S7 |
| Multi-class with important minority classes | macro F1 or balanced accuracy | per-class metrics, confusion matrix | S4 |
| Regression, large errors costly | RMSE | MAE, error distribution | S9 |
| Regression, typical error in target units | MAE | RMSE, MedAE | S9 |
| Relative error, targets far from zero | MAPE or sMAPE | MAE, the zero-target rule | S10 |
| Agreement between raters | Cohen's kappa | observed agreement, marginal distribution | S8 |

The purpose of each primary metric is specified in Table 16. The extra information from complementary metrics is also listed. This table provides an initial basis for metric choice. An application-specific cost or utility analysis is still required [28].

### 4.4. Comparison with existing guidance

Existing guidance varies in scope and in access to inputs for repeat calculations. The present guide is compared with closely related works in Table 17.

Table 17. Scope of closely related guidance and the present guide. A descriptive comparison is provided. Superiority cannot be inferred.

| Work | Scope | Examples | Tool | Consistency testing |
|---|---|---|---|---|
| Sokolova and Lapalme [5] | classification measures and their invariances | analytical | none | no |
| Chicco and Jurman [12] | binary classification; MCC, F1 and accuracy | confusion-matrix use cases | none | no |
| Hicks and colleagues [41] | binary metrics in medical artificial intelligence | published gastroenterology studies | MediMetrics web calculator | metrics derived from reported subsets |
| Maier-Hein and colleagues; Reinke and colleagues [4, 42] | image-analysis validation | illustrated pitfalls | recommendation framework | no |
| Fazekas and Kovács; Kovács and Fazekas [44, 45] | reported performance scores | published studies | mlscorecheck | yes |
| This guide | classification, probability, regression and inference | 15 locked scenarios with released inputs | scripts and checklist | four outcome classes using mlscorecheck |

Metric explanations, calculators and consistency tests are available in separate works, as shown in Table 17. These works have different aims and cover different domains. Superiority cannot therefore be inferred from this comparison.

Fixed inputs across metric families are provided in this guide. A goal-based metric table and a rule for reports of consistency checks are also included. The tests of Fazekas and Kovács are retained as the basis for these checks [44]. Metric choice is guided by the application problem, in line with Metrics Reloaded [4].

### 4.5. Reporting guide

A short set of report items was derived from the scenarios. Table 18 links each item to the scenario that motivates it.

Table 18. Checklist of report items derived from the scenarios.

| Item | What to report | Scenario |
|---|---|---|
| Test set | number of cases per class and the prevalence | S1, S2 |
| Positive class | the class named positive for every binary metric | S3 |
| Imbalance | balanced accuracy, sensitivity, specificity and MCC beside accuracy | S1 |
| Multi-class | the averaging rule and per-class values | S4 |
| Agreement | kappa with the marginal distribution | S8 |
| Threshold | the threshold, the rule that chose it, its data split and any cost ratio | S6 |
| Ranking | ROC-AUC with average precision and its baseline | S5 |
| Probabilities | a proper scoring rule, calibration slope and intercept, and the ECE binning | S7 |

| Item | What to report | Scenario |
|---|---|---|
| Regression | MAE and RMSE together, an error plot, the zero-target rule and the $R^2$ reference | S9–S11 |
| Uncertainty | n, a 95% interval with its method and the resampling unit | S12 |
| Evaluation unit | segment, participant or fold level, and pooled or averaged aggregation | S13, S14 |
| Recomputability | the confusion matrix or a prediction file, and the reporting precision of each value | S14 |
| Software | library, version and options such as zero_division and labels | S15 |

Table 18 lists the minimum information needed to interpret and recompute a reported metric. The checklist is intended for use alongside domain standards such as TRIPOD+AI [3].

### 4.6. Advantages of the proposed guide

The main practical advantages are listed below.

- Every input is released as a file, and every value can be recomputed with independent software.
- One notation covers confusion-matrix, ranking, probability, regression and inferential metrics.
- The prevalence coding of Figure 2 shows which metrics are invariant to a change in class mix at fixed class-wise rates.
- The four outcome classes of Table 4, together with the stated reporting precision, prevent a non-assessable check from being reported as an error.

### 4.7. Limitations and future work

The main limitations are listed below.

- The scenarios are synthetic and do not estimate how often each mechanism affects published results.
- The literature audit and the real-data demonstration were not executed and are specified only in Appendix D.
- Random inputs were drawn from one seed per scenario, and variation across seeds was not quantified.
- The cluster bootstrap in S13 rests on 10 constructed participants, so its intervals are themselves imprecise.
- Multi-label classification, segmentation, object detection and generative-model evaluation were outside the scope.
- Decision-analytic measures such as net benefit were cited but not examined in a scenario [28].
- The comparison with existing guidance in Table 17 is descriptive and was not based on a systematic search.

The next evaluation should execute the audit of Appendix D.1 on 200 articles with coding form version 1.0. Its consistency rate should be reported only over assessable articles, and inter-rater agreement should be measured on a subsample coded by a second person. The real-data demonstration of Appendix D.2 should then test whether the mechanisms of S2, S5, S7 and S9 change model rankings on four public tasks.

### 4.8. Potential applications

The guide can serve three groups. Authors can use Tables 16 and 18 when they choose and report metrics. Reported values can be checked by reviewers with mlscorecheck and classified with Table 4. The fixed inputs can also be used for teaching. Students can alter one factor and recalculate all metrics.

## 5. Conclusions

A methodological tutorial on performance metrics was developed. Its inputs were released and verified. A prevalence-coded metric map, 15 controlled scenarios and a goal-based metric table are provided. A four-class framework is used for consistency checks. Established metrics and tests are retained. The new elements are the fixed inputs, verification code and rules for metric reports.

An accuracy of 0.990 was obtained for a rule with no positive predictions. Precision fell from 0.889 to 0.008 solely due to prevalence. The same ROC-AUC of 0.825 was obtained for three probability variants. Their calibration slopes ranged from 0.52 to 2.08. All 205 checks passed. Of these, 202 were independent.

Individual mechanisms were isolated by design. Their frequency in published studies was not measured. The literature audit and real-data demonstration in Appendix D are proposed as the next steps. These protocols must be completed before conclusions about published practice can be drawn. The present guide can be used for metric choice and reporting within this scope.

## Declarations

**Ethical approval and informed consent:** Not applicable. Only constructed data were used. No human participants or animals were involved.

**Data availability:** The locked scenario inputs, results files, verification log (verify_checks.csv) and scenario specification are included in the archive ancillary_files_v2.zip. The archive is provided as ancillary files with the arXiv submission. It is listed in the anc/ directory of the arXiv record.

**Code availability:** The same archive contains the metric implementation, scenario builder, verification script, figure scripts, manuscript build scripts, an environment specification (environment.txt) and the run commands (README.md).

**Funding:** This research received no external funding.

**Competing interests:** The authors declare no competing interests.

**CRediT authorship contribution statement: Mehmet Baygin:** Conceptualization, Methodology, Software, Formal analysis, Writing – original draft. **Sengul Dogan:** Validation, Writing – review and editing. **Turker Tuncer:** Methodology, Supervision, Writing – review and editing.

## Appendix A. Locked scenario inputs

This appendix lists the locked input files of the scenarios. Appendix Table A1 gives each file with its number of rows and its MD5 checksum, so that any copy can be checked against the version used here.

Table A1. Locked scenario inputs. The header line is excluded from row counts. Seed: random seed of generated inputs; fixed values are typed into the builder script.

| File | Rows | MD5 checksum | Seed |
|---|---|---|---|
| S4_confusion_matrix.csv | 3 | 271c0bdba80ff6127d8cb044aaf57b2c | fixed values |
| S5_scores_balanced.csv | 200 | 6ec7bf6e50b36178e315d65bdbcd793a | 20261102 |
| S5_scores_prev001.csv | 10,000 | 1f983dbe4449d9f635a99091e24f477f | 20261102 |
| S6_test.csv | 2,000 | fb1828efe6975430db94f796a7ca53ff | 20261108 |
| S6_validation.csv | 2,000 | b21f4b0d3662a0e46e44a308920a874d | 20261108 |
| S7_logits.csv | 5,000 | c52c4aca050cb901b81eaf25ca56f0b9 | 20261109 |
| S9_regression.csv | 100 | f492c4e8559764446cea77cebb2a2ebe | 20261111 |
| S10_near_zero.csv | 10 | dec629f50d9cb178d007dcdb78c02a0c | fixed values |
| S11_negative_r2.csv | 21 | 87929f6a37db4b9a525ddf5d2588b745 | fixed values |
| S13_participants.csv | 10 | 8bef723eab1d8a55ef021f85548d2e00 | fixed values |
| S15_coarse_scores.csv | 10,000 | 3b40ef335adb5d1b7709b73183e6bfdb | fixed values |

Appendix Table A1 fixes the inputs behind every number in Sections 3.1–3.4. A file whose checksum differs from the listed value was not used in this study.

## Appendix B. Verification summary

Independent verification is summarized in this appendix. The number of checks and the independent implementation for each scenario group are listed in Appendix Table B1.

Table B1. Verification checks by scenario. Kind defines the verification method. Independent library: a value is recomputed with a function from another library. Monte Carlo: a library function is applied to random resamples or permutations. Results are compared within a Monte Carlo tolerance. Independent re-implementation: the formula is coded separately. Internal: the same calculation is repeated from the input file. All checks passed.

| Scenario | Checks | Kind | Reference implementation | Tolerance |
|---|---|---|---|---|
| S1 | 12 | independent library | scikit-learn | $10^{-9}$ |
| S2 | 42 | independent library | scikit-learn | $10^{-9}$ |
| S3 | 49 | independent library | scikit-learn | $10^{-9}$ |
| S4 | 6 | independent library | scikit-learn | $10^{-9}$ |
| S5 | 4 | independent library | scikit-learn | $10^{-9}$ |
| S5 | 1 | independent library (Monte Carlo) | scikit-learn average_precision_score over random permutations (20,000 balanced, 2,000 low-prevalence; seed 7) | 0.002 |
| S5 | 1 | independent library (Monte Carlo) | scikit-learn average_precision_score over random permutations (20,000 balanced, 2,000 low-prevalence; seed 7) | 0.0005 |
| S5 | 1 | independent re-implementation | enumeration of both rankings | $10^{-12}$ |
| S6 | 11 | independent library | scikit-learn | $10^{-9}$ |
| S6 | 5 | independent re-implementation | scikit-learn confusion_matrix counts + cost formula | $10^{-9}$ |
| S7 | 9 | independent library | scikit-learn | $10^{-9}$ |
| S7 | 3 | independent library | scikit-learn LogisticRegression | $10^{-4}$ |
| S7 | 9 | independent re-implementation | scikit-learn calibration_curve bin statistics + bin weighting | $10^{-6}$ |
| S7 | 3 | independent re-implementation | SciPy optimizer on separately coded likelihood | $10^{-5}$ |

| Scenario | Checks | Kind | Reference implementation | Tolerance |
|---|---|---|---|---|
| S8 | 14 | independent library | scikit-learn | $10^{-9}$ |
| S9 | 8 | independent library | scikit-learn | $10^{-9}$ |
| S10 | 1 | independent library | scikit-learn | $10^{-9}$ |
| S11 | 3 | independent library | scikit-learn | $10^{-9}$ |
| S12 | 6 | independent library | statsmodels proportion_confint | $10^{-9}$ |
| S13 | 2 | independent library | statsmodels proportion_confint | $10^{-9}$ |
| S13 | 4 | independent library (Monte Carlo) | SciPy bootstrap over participants (percentile; 10,000 resamples; seed 11) | 0.02 |
| S13 | 3 | internal | pandas re-derivation from the input file | $10^{-9}$ |
| S14 | 7 | independent library | mlscorecheck | $10^{-9}$ |
| S15 | 1 | independent library | scikit-learn average_precision_score | $10^{-12}$ |

All checks passed, as summarized in Appendix Table B1. The scenario calculations were confirmed by agreement with independent libraries. The scenario design itself was not validated by this agreement.

**Appendix C. Generating models and fold counts**

The random inputs and fold example can be regenerated from the details in this appendix without the code archive. The model, seed and size for each random input are listed in Appendix Table C1.

Table C1. Models used to generate random scenario inputs. N(μ, σ²) = normal distribution; U(a, b) = uniform distribution; sigmoid(z) = 1 / (1 + $e^{-z}$). Inputs were drawn once with the NumPy default_rng generator and saved. The saved inputs are reused without regeneration.

| Scenario | Seed | Generating model | Size |
|---|---|---|---|
| S5 | 20261102 | positive scores N(1.5, 1); negative scores N(0, 1); the prevalence-0.010 set repeats each negative score 99 times | 100 + 100; 100 + 9,900 |
| S6 | 20261108 | logit z: positives N(1.0, 1.2²), negatives N(−2.0, 1.2²); score = sigmoid(z); validation set drawn first, then test set from the same generator | 2 × 2,000 (200 positive each) |
| S7 | 20261109 | logit z ~ N(−1, 1.5²); label y ~ Bernoulli(sigmoid(z)), so sigmoid(z) is the true event probability; variants sigmoid(a · z) | 5,000 |
| S9 | 20261111 | targets U(20, 80) rounded to 0.1; error signs alternate; model B has errors of 30 at indices 7, 23, 46, 68 and 91 (0-based) and 0.5 elsewhere | 100 |
| S13 | 20261115 | cluster bootstrap: 10 participants resampled with replacement | 10,000 resamples |

The distributions for S5, S6, S7, S9 and the S13 bootstrap are specified in Appendix Table C1. The same generator, seed and order of random draws must be used to reproduce the released inputs.

Counts and fold-wise scores for all five folds of S14 are listed in Appendix Table C2.

Table C2. Fold counts for the five-fold example in S14. Counts are listed as TP / FN / FP / TN. Each fold contains 100 cases.

| Fold | Counts | P | N | Acc | TPR | TNR |
|---|---|---|---|---|---|---|
| 1 | 18 / 2 / 10 / 70 | 20 | 80 | 0.880 | 0.900 | 0.875 |
| 2 | 9 / 1 / 6 / 84 | 10 | 90 | 0.930 | 0.900 | 0.933 |

| Fold | Counts | P | N | Acc | TPR | TNR |
|---|---|---|---|---|---|---|
| 3 | 25 / 5 / 7 / 63 | 30 | 70 | 0.880 | 0.833 | 0.900 |
| 4 | 12 / 3 / 9 / 76 | 15 | 85 | 0.880 | 0.800 | 0.894 |
| 5 | 20 / 5 / 8 / 67 | 25 | 75 | 0.870 | 0.800 | 0.893 |
| Pooled | 84 / 16 / 40 / 360 | 100 | 400 | 0.888 | 0.840 | 0.900 |
| Mean of folds | not applicable | | | 0.888 | 0.847 | 0.899 |

Equal-sized folds with different prevalence are compared in Appendix Table C2. Fold means differ from pooled sensitivity and specificity. Mean and pooled accuracy agree because the same number of cases is included in every fold.

## Appendix D. Pre-specified protocols for empirical extension

Two empirical extensions are specified in this appendix. Neither was carried out in the present study. Version 1.0 of both protocols is frozen with this paper; any later change is to be reported as a deviation.

### D.1. Literature audit

The planned audit will estimate the frequency of reports with the information required in Sections 3.1–3.4.

- Sources: OpenAlex works of type article published from 1 January 2022 to 31 December 2025 with open full text, checked against Crossref.
- Classification query: title and abstract search for (“machine learning” OR “deep learning”) AND (classification OR classifier) AND (accuracy OR sensitivity OR F1 OR AUC).
- Regression query: title and abstract search for (“machine learning” OR “deep learning”) AND regression AND (RMSE OR “mean absolute error” OR “R2”).
- Strata and sample: 150 classification articles (37 or 38 per publication year) and 50 regression articles (12 or 13 per year), drawn with the OpenAlex sample parameter and seed 20261102 within each stratum.
- Exclusions: reviews, editorials and protocols; segmentation, detection and generative-model studies; studies without a numerical test-set result; and duplicates. Excluded records are replaced by the next sampled record of the same stratum.
- Result selection: the headline result of the model proposed by the authors on the first held-out test set; when no held-out set exists, the first cross-validation result reported in the results section.
- Pilot: the first 20 articles test coding form version 1.0 (Appendix Table D1); any change creates version 1.1 before the main sample is coded.
- Reliability: a second person codes a random 20% of articles independently, and inter-rater agreement is reported with Cohen’s kappa. The first coder recodes another random 20% after at least two weeks, and this intra-rater agreement is reported separately and labelled as such.

- Consistency checks: every check is run with mlscorecheck under the model stated by the article and classified with Table 4. Each flagged inconsistency is reviewed again.
- Reporting: proportions are reported with Wilson intervals. Only assessable articles are included in the denominator of each consistency rate. Relevant items from PRISMA 2020 are applied [49].

Table D1. Coding form version 1.0 for the pre-specified literature audit. Each field is assigned one of the listed values. Separate codes are used for not reported, not applicable, not assessable and inconsistent.

| Field | Values |
|---|---|
| Task type | binary; multi-class; regression |
| Test-set size and class distribution | counts; not reported |
| Metrics reported | list of metric names |
| Averaging rule | micro; macro; weighted; not reported; not applicable |
| Positive class named | yes; no; not applicable |
| Threshold and its selection rule | value and rule; not reported; not applicable |
| Evaluation unit | instance; participant; fold mean; not reported |
| Confusion matrix or prediction file | available; not available |
| Uncertainty | interval method and resampling unit; not reported |
| Reporting precision | number of decimals per metric |
| Consistency check | consistent; consistent within rounding; inconsistent; not assessable |

The audit fields and codes are defined in Appendix Table D1. “Not reported”, “not applicable”, “not assessable” and “inconsistent” are recorded separately. An absent confusion matrix alone is not evidence of an incorrect result.

### D.2. Real-data demonstration

The planned demonstration will test whether model order on real data is altered by the scenario mechanisms.

- Tasks: four public-data tasks are selected. These comprise moderately balanced and strongly imbalanced binary tasks, a multi-class image task, and a regression task with targets near zero.
- Models: at least six models are evaluated per task. A convolutional network is included for the image task.
- Design: thresholds and calibration are tuned only on validation data. All metrics are computed from the same test predictions.

- Analysis: model orders from each metric are compared with Kendall's $\tau_b$. A bootstrap distribution is obtained over test cases. Low agreement is interpreted in relation to a predefined application goal from Table 16. It is not treated as evidence of a failed metric.